\documentclass[conference,compsoc]{IEEEtran}

\usepackage{cite}

\usepackage{tikz}
\usepackage{amsmath,amssymb,amsfonts}
\usepackage{algorithmic}
\usepackage{parskip}
\usepackage{enumitem}
\usepackage{multirow}
\usepackage{array}
\usepackage{cleveref}

\usepackage{tabularx, nicematrix}
\usepackage{booktabs}
\usepackage{pbox}
\usepackage{graphicx}
\usepackage{adjustbox}
\usepackage{flushend}
\usepackage{vcell}

\newcommand\approach{DeepBaR}

\newcommand\ASRa{$\text{ASR}_{> \tau}$}
\newcommand\ASRb{$\text{ASR}_{\leq \tau}$}
\newcommand\ASR{ASR}

\makeatletter 
\newcommand{\linebreakand}{%
  \end{@IEEEauthorhalign}
  \hfill\mbox{}\par
  \mbox{}\hfill\begin{@IEEEauthorhalign}
}
\makeatother 

\begin{document}

\title{\approach{}: Fault backdoor attack on deep neural network layers}

\author{\IEEEauthorblockN{Camilo A. Martínez-Mejía}
\IEEEauthorblockA{Independent Researcher \\
Bogotá, Colombia\\
E-mail: camilo.csmath@gmail.com}
\and
\IEEEauthorblockN{Jesus Solano}
\IEEEauthorblockA{ETH Zürich\\
Zürich, Switzerland\\
jesus.solano@inf.ethz.ch}
\and
\IEEEauthorblockN{Jakub Breier}
\IEEEauthorblockA{TTControl GmbH\\
Vienna, Austria\\
E-mail: jbreier@jbreier.com}
\linebreakand 
\IEEEauthorblockN{Dominik Bucko}
\IEEEauthorblockA{Deutsche Telekom Cloud Services\\
Bratislava, Slovakia\\
bucko.dominik@protonmail.com}
\and
\IEEEauthorblockN{Xiaolu Hou}
\IEEEauthorblockA{Slovak University of Technology\\
Bratislava, Slovakia\\
houxiaolu.email@gmail.com}}

\maketitle

\begin{abstract}
Machine Learning using neural networks has received prominent attention recently because of its success in solving a wide variety of computational tasks, in particular in the field of computer vision. However, several works have drawn attention to potential security risks involved with the training and implementation of such networks. In this work, we introduce \approach{}, a novel approach that implants backdoors on neural networks by faulting their behavior at training, especially during fine-tuning. Our technique aims to generate adversarial samples by optimizing a custom loss function that mimics the implanted backdoors while adding an almost non-visible trigger in the image. We attack three popular convolutional neural network architectures and show that \approach{} attacks have a success rate of up to 98.30\%. Furthermore, \approach{} does not significantly affect the accuracy of the attacked networks after deployment when non-malicious inputs are given. Remarkably, \approach{} allows attackers to choose an input that looks similar to a given class, from a human perspective, but that will be classified as belonging to an arbitrary target class. 
\end{abstract}

\IEEEpeerreviewmaketitle

\section{Introduction}
In recent years, machine learning using neural networks has seen significant progress \cite{jordan2015machine}. For instance, classification tasks in computer vision have reached high levels of accuracy, and are fundamental for autonomous driving among others \cite{liu2020computing}. Naturally, concerns about the security and safety of such solutions have been raised and studied \cite{shafaei2018uncertainty}. A natural concern arising in this context is: Can inputs to computer vision classifiers (images), which would be classified correctly by a human observer be crafted by adversaries such that they will be misclassified by machine learning algorithms? A motivation for an attacker to perform such attacks would be to maliciously cause an autonomous vehicle to misbehave or to bypass an authentication mechanism based on face recognition among others \cite{deng2020analysis}. Such attacks have been shown to be possible using \emph{adversarial learning} \cite{lowd2005adversarial}, a technique that exploits the fact that boundaries between classes in a machine learning classifier are easy to cross by small perturbations to inputs. 

However, a much harder problem is to achieve \emph{targeted} adversarial attacks, where an adversary produces inputs that look (to humans) as if they would belong to a given class but will be misclassified as an arbitrary target class \cite{ilyas2018black}. 
The difficulty of targeted attacks comes from the need to navigate the model’s decision boundaries in high-dimensional space to find a specific adversarial example that leads to a specific incorrect classification. This requires a more sophisticated understanding of the model’s internal workings and often requires more computational resources compared to non-targeted attacks where the goal is to find any point that crosses the model’s decision boundary.
In previous work \cite{breier2022foobar}, it has been shown possible to achieve a higher degree of success rate by introducing backdoors to neural networks under training by using fault injection attacks \cite{breier2022practical}. Although promising, this approach had several drawbacks. First, its success depended on finding the solution to a constraint-solving problem, which was shown to work only in simple network architectures where fault injections were performed at superficial layers. This technique for instance cannot be directly applied to modern convolutional neural networks such as VGG-19~\cite{simonyan2014very}.

In this work, we show a novel technique to insert backdoors in complex convolutional network architectures while achieving a high attack success rate, called \approach{}. At the core of our technique is a faulting attack similar to the one presented in \cite{breier2022foobar} but which does not rely on constraint solving to find the fooling images. Instead, we propose a strategy that relies on generating adversarial samples by optimizing custom loss functions that mimic the fault attacks performed during training while keeping the image similarity, so attacks (i.e. changes in the image) are hardly perceptible by humans. We evaluate our approach on highly popular convolutional neural networks such as VGG-19~\cite{simonyan2014very}, ResNet-50~\cite{he2016deep}, and DenseNet-121~\cite{huang2017densely}. As a result, we obtain high attack success rates, of up to $98.30\%$ for VGG-19, $97.94\%$ for ResNet-50, and $88.94\%$ for DenseNet-121. 

When comparing our approach to traditional targeted adversarial examples~\cite{weng2023comparative}, it provides several advantages.
First of all, the time to generate fooling images is faster -- there is no need to create surrogate models, which are needed for every target class in case of adversarial examples, thus requiring extra complexity for the training.
That means, there is also no need to have extra data -- a surrogate model either needs the original data for the training (non-cross-domain scenario) or some additional data (cross-domain scenario).
Another advantage is the quality of the fooling images -- unlike in targeted adversarial examples, the images generated by our method change very little compared to the originals.
Finally, we also showed that by performing traditional adversarial training, the \ASR{} of \approach{}'s adversarial samples drops significantly for all the datasets and architectures studied.

\begin{figure*}[htbp]
    \centering
    \textbf{Original images} \\
        \vspace{0.5em}
    \begin{minipage}{0.19\textwidth}
        \centering
        \includegraphics[width=\textwidth]{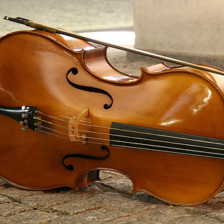}
    \end{minipage}
    \begin{minipage}{0.19\textwidth}
        \centering
        \includegraphics[width=\textwidth]{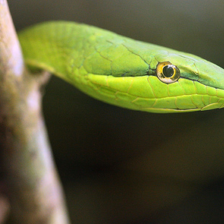}
    \end{minipage}
    \begin{minipage}{0.19\textwidth}
        \centering
        \includegraphics[width=\textwidth]{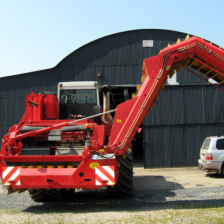}
    \end{minipage}
    \begin{minipage}{0.19\textwidth}
        \centering
        \includegraphics[width=\textwidth]{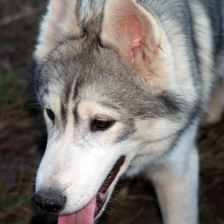}
    \end{minipage}
    \begin{minipage}{0.19\textwidth}
        \centering
        \includegraphics[width=\textwidth]{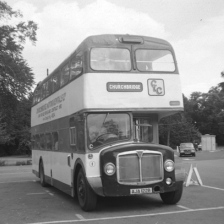}
    \end{minipage}

    \vspace{0.5em} 
    
    \centering
    \textbf{\approach{} Adversarial Samples for VGG-19} \\
        \vspace{0.5em}
    \begin{minipage}{0.19\textwidth}
        \centering
        \includegraphics[width=\textwidth]{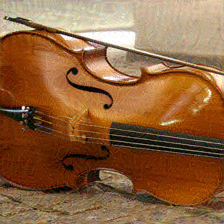}
    \end{minipage}
    \begin{minipage}{0.19\textwidth}
        \centering
        \includegraphics[width=\textwidth]{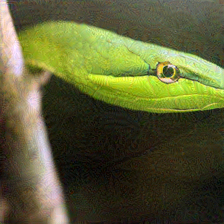}
    \end{minipage}
    \begin{minipage}{0.19\textwidth}
        \centering
        \includegraphics[width=\textwidth]{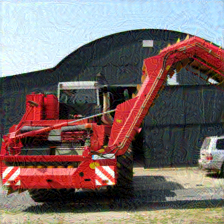}
    \end{minipage}
    \begin{minipage}{0.19\textwidth}
        \centering
        \includegraphics[width=\textwidth]{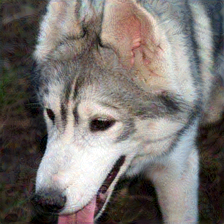}
    \end{minipage}
    \begin{minipage}{0.19\textwidth}
        \centering
        \includegraphics[width=\textwidth]{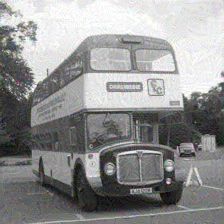}
    \end{minipage}

    \centering
    \vspace{0.5em}
    \textbf{\approach{} Adversarial Samples for ResNet-50} \\
    \vspace{0.5em}
    \begin{minipage}{0.19\textwidth}
    \centering
    \includegraphics[width=\textwidth]{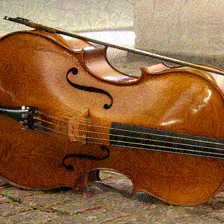}
    \end{minipage}
    \begin{minipage}{0.19\textwidth}
        \centering
        \includegraphics[width=\textwidth]{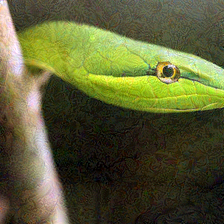}
    \end{minipage}
    \begin{minipage}{0.19\textwidth}
        \centering
        \includegraphics[width=\textwidth]{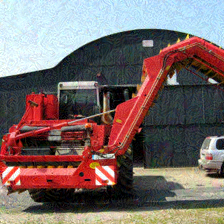}
    \end{minipage}
    \begin{minipage}{0.19\textwidth}
        \centering
        \includegraphics[width=\textwidth]{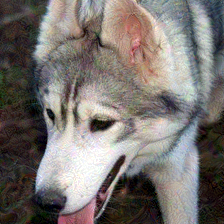}
    \end{minipage}
    \begin{minipage}{0.19\textwidth}
        \centering
        \includegraphics[width=\textwidth]{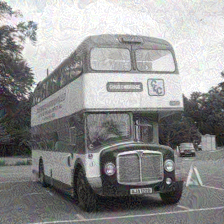}
    \end{minipage}

    \centering
    \vspace{0.5em}
    \textbf{\approach{} Adversarial Samples for DenseNet-121} \\
    \vspace{0.5em}
        \begin{minipage}{0.19\textwidth}
    \centering
    \includegraphics[width=\textwidth]{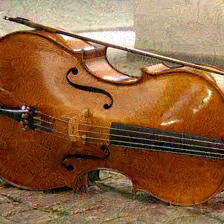}
    \end{minipage}
    \begin{minipage}{0.19\textwidth}
        \centering
        \includegraphics[width=\textwidth]{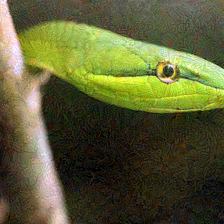}
    \end{minipage}
    \begin{minipage}{0.19\textwidth}
        \centering
        \includegraphics[width=\textwidth]{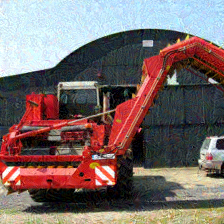}
    \end{minipage}
    \begin{minipage}{0.19\textwidth}
        \centering
        \includegraphics[width=\textwidth]{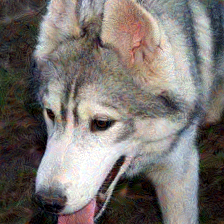}
    \end{minipage}
    \begin{minipage}{0.19\textwidth}
        \centering
        \includegraphics[width=\textwidth]{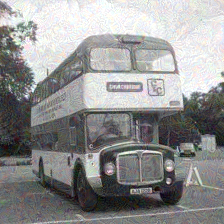}
    \end{minipage}
    
    \caption{Graphical representation of original testing images and adversarial samples generated using \approach{}. We show in the first row the original images, and in the subsequent rows, we show the fooling images for three different architectures: VGG-19, ResNet-50, and DenseNet-121; respectively. For each architecture, we depict a block of 5 images that are classified as \textit{Great Grey Owl} regardless of the input.}
    \label{fig:images_vs_fooling_images}
\end{figure*}

In sum, our contributions are:

\begin{itemize}
    \item We propose, for the first time, a fault-based attack applied during the training, particularly in the context of fine-tuning, for complex convolutional neural network architectures that plants a backdoor in the model allowing a trigger-based targeted misclassification.
    \item Our attack can be easily applied by using simple fault injection techniques producing instruction skips, such as clock/voltage glitch, thus removing the need for a complicated Rowhammer-based attack on memories.
    \item Generation of fooling images does not require training of surrogate models, lowering the attack complexity.
    \item The accuracy of the backdoored models changes very little compared to the original models. Particularly, \approach{} reduces the performance of the downstream model by up to 0.8\% on average.
    \item We evaluate our approach and show it has higher attack success rates than the current state-of-the-art weight-oriented approach, DeepVenom~\cite{yao2024deepvenom}.
    \item We propose adversarial training as a countermeasure to our approach showing that after re-training with fooled images generated with \approach{} the \ASR{} drops to roughly 5\% regardless the architecture.
\end{itemize}

The rest of the paper is organized as follows. Section~\ref{sec:background} recaps fundamental machine learning and fault attack concepts. Section~\ref{sec:approach} describes our approach and faulting strategy. Section \ref{sec:evaluation} describes the experimental setup, datasets considered for the evaluation, and the evaluation protocol. Section \ref{sec:results} presents the results of the evaluation for in-domain and out-of-domain data. Section \ref{sec:countermeasures} shows a set of strategies to defend against \approach{}. Finally, Section \ref{sec:related} shows a comprehensive summary of the state-of-the-art approaches to adversarial and fault-based backdoor attacks.

\section{Background and preliminaries}
\label{sec:background}
In this section, we will first give an introduction to neural networks (Section~\ref{sec:anns}) and fault attacks on neural networks (Section~\ref{sec:fas}).
Section~\ref{sec:ssim} provides an overview of the Structural Similarity Index (SSIM). 
Then in Section~\ref{sec:terms} we present the terminologies that will be used for the rest of this paper.
Finally, Section~\ref{sec:attackmetric} formalizes fooling backdoor attacks.

\subsection{Artificial neural networks}
\label{sec:anns}
Artificial neural networks (ANNs) are a subset of supervised machine learning algorithms that use a network of interconnected artificial neurons to simulate the function of a human brain.
Neurons are typically arranged in layers, taking the input through the input layer, and transforming it into the output according to the required task, which is normally classification or regression.
If there is at least one other layer between the input and the output layers (a \textit{hidden} layer), we talk about deep neural networks (DNNs) which are at the heart of deep learning.
Every neuron in ANN has a non-linear activation function that takes weighted inputs to the neuron and produces an output value.
In our work, we are interested in networks based on ReLU (rectifier linear unit) activation function~\cite{agarap2018deep}.
ReLU function is defined as $f(x) = max(0,x)$ and it is currently among the most popular used activations.

ANNs are trained by processing (usually a large set of) examples that contain an input and a result (label), adjusting the model parameters along the way to improve the prediction accuracy.
The training algorithm that performs this step is called \textit{backpropagation}~\cite{gurney2018introduction}.

\noindent
\textbf{Convolutional neural networks}
Convolutional neural networks (CNNs) were designed to improve image classification tasks by recognizing specific patterns in the (mostly 2-dimensional) input data.
CNNs are ANNs that utilize a form of feature extraction within their convolutional layers~\cite{albawi2017understanding}.
Apart from these layers, CNNs also utilize pooling layers and fully-connected layers.
Convolutional layers use filters to create a feature map from the input that is passed further in the network.
Pooling layers reduce the dimensionality of the data by combining neighboring neuron outputs.
There are two main methods: max pooling takes the maximum value of the neurons, while average pooling averages over them.
Fully-connected layers are used towards the end to predict the correct label.

\subsection{Fault attacks on neural networks}
\label{sec:fas}
Fault attacks are active hardware-level attacks that target the device executing the implementation.
Originally, they were proposed against cryptographic algorithms, allowing very efficient key recovery attacks~\cite{barenghi2012fault}.
They can either be performed with physical access to the device, for example by means of clock/voltage glitching, electromagnetic pulses, lasers~\cite{breier2022practical}, or remotely by using techniques such as Rowhammer~\cite{kim2014flipping}, CLKSCREW~\cite{tang2017clkscrew} or VoltJockey~\cite{qiu2019voltjockey}.
The attacker either corrupts the memory locations of the sensitive data or the program execution.
In the first case, corruption can lead to bit-flips, bit sets/resets, or random byte faults~\cite{breier2022practical}.
In the second case, the attacker can alter the program execution by skipping~\cite{breier2015laser} or changing~\cite{kumar2017practical} some instructions.

The first fault attack proposal targeting neural network models was published in 2017~\cite{liu2017fault}, utilizing bit-flips to cause misclassification.
The first practical attack followed a year after and it used instruction skips caused by a laser to change the output of executed activation functions~\cite{breier2018practical}.
After that, the main body of work used Rowhammer faults to flip bits in DRAM memories~\cite{hong2019terminal,he2020defending,rakin2021t,bai2021targeted}.
The Rowhammer-based methods were termed \textit{adversarial weight attacks}, and apart from~\cite{hong2019terminal}, they all targeted quantized neural networks -- models that use low-precision data representation, typically with a word-length of 8 or 4 bits instead of a traditional floating point representation (IEEE 754).
Apart from misclassification, other attack vectors have been investigated, such as model extraction~\cite{breier2021sniff,rakin2022deepsteal}, trojan insertion~\cite{rakin2020tbt}, and backdoor injection~\cite{breier2022foobar}.
In this work, we aim at backdoor injection as well, but unlike in~\cite{breier2022foobar}, our method can scale up to large real-world models, thus bringing this attack vector to the practical realm.\\

\begin{figure}
    \centering
    \includegraphics[width=0.8\linewidth]{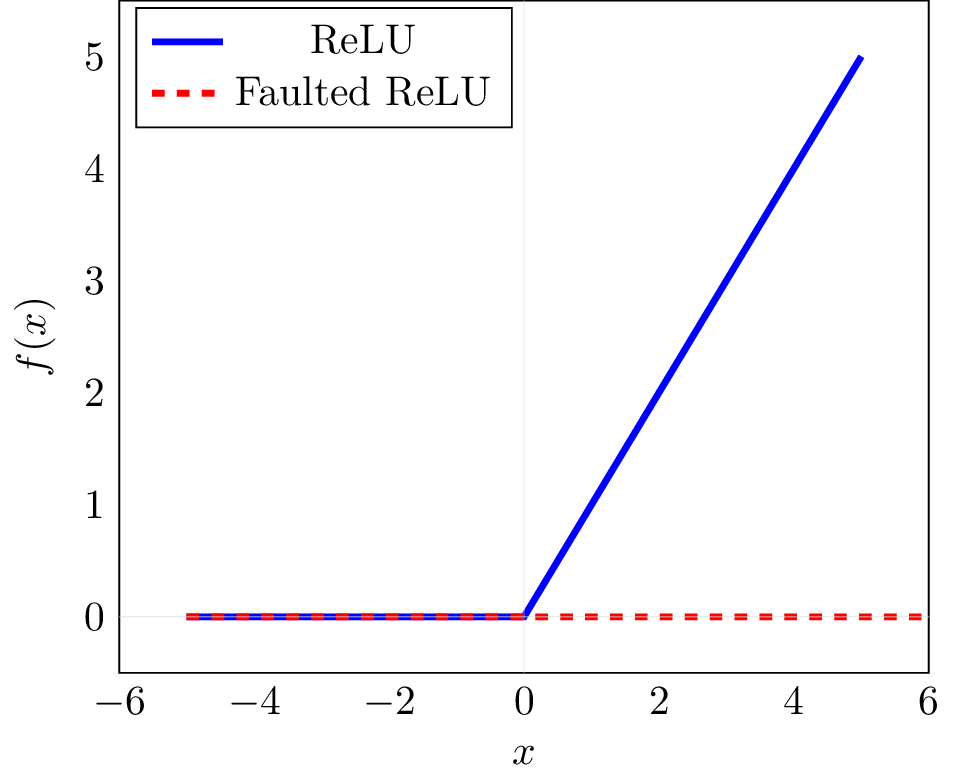}
    \caption{Illustration of the ReLU-skip attack.}
    \label{fig:reluskip}
\end{figure}

\noindent
\textbf{ReLU-skip attack}
At the core of this work, there is a fault model that was shown to be practical in~\cite{breier2018practical}.
The main idea is to skip a jump instruction during the ReLU execution to always force the function output to be zero, as illustrated in Figure~\ref{fig:reluskip}.
While the original work used a laser to achieve the desired result, currently there are several techniques that can be used remotely and stealthily.
For example, CLKSCREW~\cite{tang2017clkscrew} and VoltJockey~\cite{qiu2019voltjockey} use energy management features in the software to exploit frequency and voltage hardware regulators to remotely skip instructions on ARM processors.
In~\cite{gross2023fpganeedle}, the authors show how voltage drop from an FPGA located on the same die as a CPU can cause instruction skips.
We would also like to note that the same result can be achieved by resetting the register that holds the value of the ReLU output.
This extends the attack to models implemented on FPGAs, as the register reset on these devices was successfully demonstrated in~\cite{courbon2015combining}.
In the following, we refer to this fault model as \textit{ReLU-skip attack}.

\subsection{Faulting assumptions}
\label{sec:faultassu}
Rowhammer attacks assume the attacker has access to the victim's memory -- by means of operating system privileges allowing writing to a specific part of the memory, surrounding the sensitive data.

Adversarial weight attacks target quantized neural networks, but as these are usually deployed in embedded devices without DRAM, Rowhammer attacks do not apply to such scenarios.

Apart from that, there is a wide body of work dedicated to Rowhammer defenses~\cite{mutlu2019rowhammer}.

On the other hand, causing instruction skips is relatively easy~\cite{breier2018practical}. In the context of \approach{}, we assume that the attacker can execute \textit{ReLU-skip attacks}, and has access to weights and implementation code of the infected model. This access allows them to leverage its methods and attributes.

\subsection{SSIM}
\label{sec:ssim}
The Structural Similarity Index (SSIM)~\cite{wang2004image} is a metric used to quantify the similarity between two images. Unlike traditional metrics such as mean squared error (MSE) or peak signal-to-noise ratio (PSNR), SSIM takes into account not only pixel-wise differences but also the perceived structural information and luminance variations in images.

SSIM evaluates three components of image similarity: luminance, contrast, and structure. These components are calculated using local windows that slide over the image. Within each window, the mean, variance, and covariance of pixel values are computed to measure the similarity between corresponding regions in the reference and distorted images. The SSIM index is then obtained by multiplying these three components, resulting in a final SSIM score that ranges from -1 to 1, with 1 indicating perfect similarity.

\subsection{Terminology}
\label{sec:terms}
Following the common terminologies in backdoor attacks on neural networks, below we detail the terms that will be used in this work:
\begin{itemize}
    \item \textit{Benign model} is the model trained without fault attacks.
    \item \textit{Infected model} is the model trained with the backdoor injected by fault attacks.
    \item \textit{Fooling input/image} is the sample input generated with our constraint solver, aimed to fool the infected model so that the output of the model is incorrect.
    \item \textit{Base pattern} is the sample used to generate fooling input from. It can be a random sample outside the problem domain.
    \item \textit{Source class} is the output of the benign model given an input.
    \item \textit{Target class} is the class the attacker would like the infected model to output given an input.
    \item \textit{Benign accuracy} is the test accuracy of the benign model.
\end{itemize}

\subsection{Fooling backdoor attack}
\label{sec:attackmetric}
Let $c_t$ denote our target class of the attack.
Faults are injected during the training of a benign model $\mathcal{M}$ and result in an infected model $\mathcal{M}'$.
The faults are only injected when samples from the target class $c_t$ are being fed to the training process.

Given any base pattern $\mathbf{x}$, with the knowledge of fault locations and effects injected during the training, the attacker constructs a fooling input $\mathbf{x}'$ from $\mathbf{x}$.
The base pattern $\mathbf{x}$ can be a legitimate sample from the test set, and can also be an out-of-the-domain sample that belongs to one of the possible classes.

An attack on $\mathbf{x}$ is said to be \textit{successful} if 
\begin{equation}\label{eq:misclassification}
    \mathcal{M}'(\mathbf{x}')=c_t.
\end{equation}

However, if the classification result of $\mathbf{x}'$ by $\mathcal{M}'$ has very low confidence, denoted CF$(\mathbf{x}')$, we would expect the victim to be suspicious about the output.
Hence, we also define a \textit{threshold} $\tau$ that the confidence should reach for a successful attack.
And we say an attack on $\mathbf{x}$ is \textit{successful above the threshold} if the misclassification from Equation~\ref{eq:misclassification} has confidence $\text{CF}(\mathbf{x}')>\tau$.
Similarly, we say an attack on $\mathbf{x}$ is \textit{successful below the threshold} if the misclassification from Equation~\ref{eq:misclassification} has confidence $\text{CF}(\mathbf{x}')\leq\tau$.

Let $c_{\mathbf{x}}$ denote the source class of $\mathbf{x}$, i.e.
\[
\mathcal{M}(\mathbf{x})=c_{\mathbf{x}}.
\]
If Equation~\ref{eq:misclassification} holds and we also have
\[
\mathcal{M}(\mathbf{x}')=c_{\mathbf{x}},
\]
then we say that the successful attack on $\mathbf{x}$ has \textit{successful validation}.

Let $S$ denote a set of $N$ base patterns
\[
S=\{\mathbf{x}_1,\mathbf{x}_2,\dots,\mathbf{x}_N\}.
\]
And let $S'$ be the set of corresponding fooling inputs
\[
S'=\{\mathbf{x}'_1,\mathbf{x}'_2,\dots,\mathbf{x}'_N\}
\]
With the above terminologies, we define the following four metrics that will be used for evaluating and presenting our attack results:
\begin{itemize}[leftmargin=*]
    \item \textbf{Attack Success Rate Above Threshold}, denoted \ASRa{}, is given by the percentage of inputs $\mathbf{x}$ in $S$ such that the attack on $\mathbf{x}$ is successful above the threshold.
    In other words, 
    \[
    \text{\ASRa{}}=\frac{\left|\{i\ |\ \mathcal{M}'(\mathbf{x}_i')=c_t,\text{CF}(\mathbf{x}')>\tau, i=1,2,\dots, N\}\right|}{N}
    \]
    \item \textbf{Attack Success Rate Below Threshold}, denoted \ASRb{}, is given by the percentage of inputs $\mathbf{x}$ in $S$ such that the attack on $\mathbf{x}$ is successful below the threshold.
    Or equivalently, 
    \[
    \text{\ASRb{}}=\frac{\left|\{i\ |\ \mathcal{M}'(\mathbf{x}_i')=c_t,\text{CF}(\mathbf{x}')\leq\tau, i=1,2,\dots,N\}\right|}{N}
    \]
    \item \textbf{Attack Success Rate}, denoted \ASR{}, is given by the percentage of $\mathbf{x}$ in $S$ such that the attack on $\mathbf{x}$ is successful.
    Namely,
    \[
    \text{\ASR{}}=\text{\ASRa{}}+\text{\ASRb{}}
    \]

\end{itemize}
\section{Our approach}
\label{sec:approach}
Our approach entails creating a backdoor in a deep neural network layer during the training or fine-tuning process, which can then be exploited during inference. We have two phases: the \textit{faulting strategy} (Section~\ref{sec:faulstrategy}) and the \textit{fooling image generation strategy} (Section~\ref{sec:foolgenstrategy}). The \textit{faulting strategy} involves the deliberate manipulation of Deep Neural Networks (DNNs) during the training phase. Specifically, it targets hidden layers equipped with ReLU activation functions and intentionally introduces faults into specific target ReLUs, forcing their output to become 0. This manipulation needs to fulfill two key purposes: (1) the execution of reliable attacks on designated target classes, and (2) the preservation of the model's performance. In contrast, the \textit{fooling image generation strategy} is designed for generating fooling images post-model deployment during the testing phase. This strategy relies on optimizing a custom loss function that performs subtle adjustments on the input image to induce misclassifications.

\begin{figure*}[h]
\centering
\includegraphics[scale=0.13]{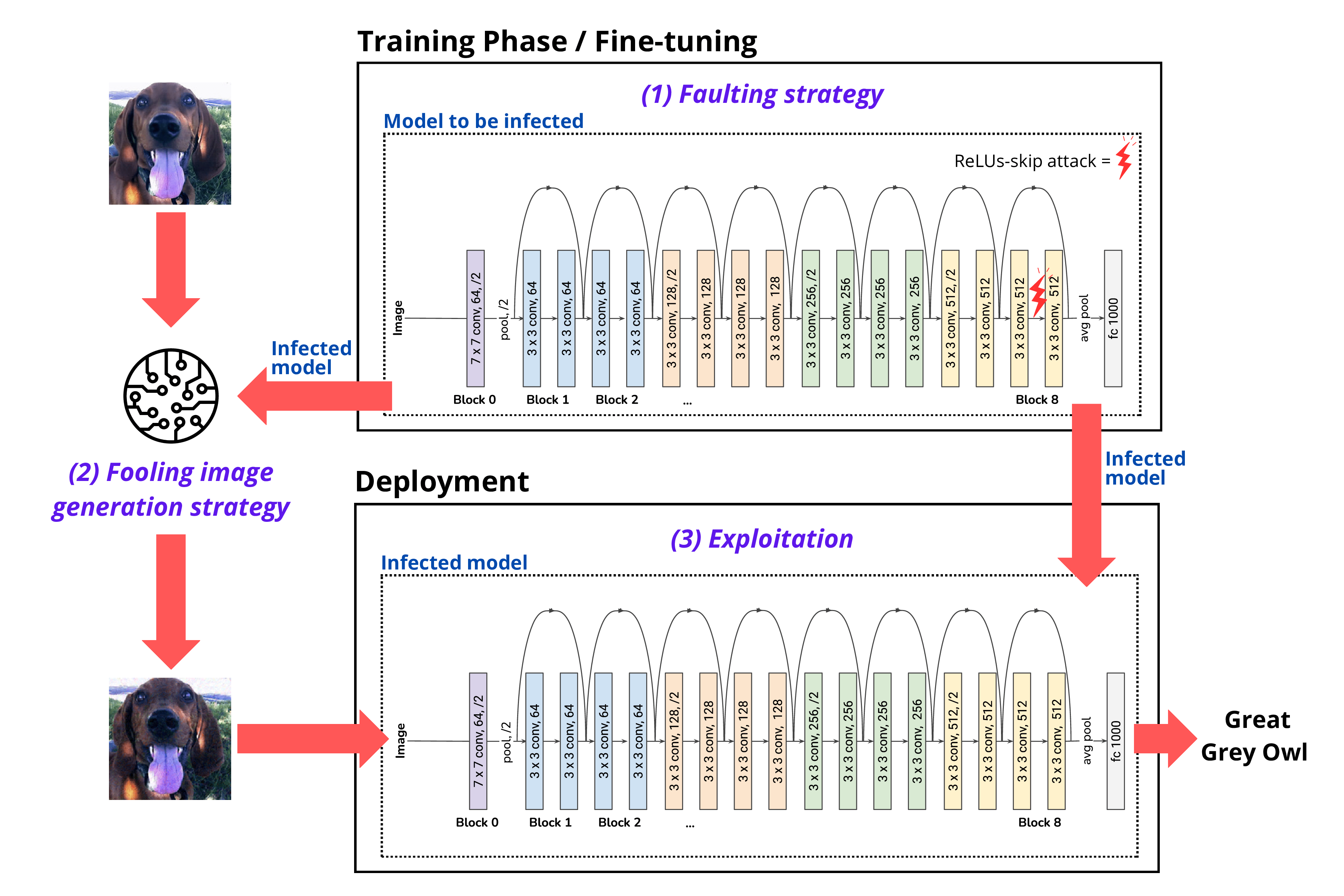}
\caption{\label{fig:attack_overview} High-level overview of the \approach{} attack: including (1) the faulting strategy, (2) the strategy for generating fooling images, and (3) the exploitation during deployment. Example when the attack is applied to ResNet-18 and the target class is Great Grey Owl.}
\end{figure*}

\subsection{Faulting strategy}
\label{sec:faulstrategy}
In this study, our primary emphasis is on Convolutional Neural Networks (CNNs) for image classification. Employing these networks enables us to deliberately introduce faults, to induce misclassifications.

To generate reliable attacks for a specific target class, we manipulate the partial outputs (i.e., the computations that follow a layer) during the training phase. This manipulation involves a physical action: executing the \textit{ReLU-skip attack} detailed in Section~\ref{sec:fas}. Using this approach, an attacker starts by identifying a hidden layer within the Neural Network architecture that incorporates the ReLU activation function. Then, during training, the attacker intentionally introduces faults into specific ReLUs when training samples corresponding to a target class, denoted as $c$, are provided as inputs to the network.
Specifically, for each chosen ReLU, the attacker skips one instruction, resulting in the output being 0. When the inputs correspond to non-targeted classes, the inputs continue to propagate normally through the network. In other words, we do not execute attacks in such cases.

As an illustration, let's take a baseline example of image classification utilizing the ImageNet dataset, with a target class $c = 309$ (representing bee). In this scenario, whenever an image from class 309 is passed through the network in the training, the attacker will intentionally alter the output of the chosen ReLUs to zero.

Given that the attacker wants to be stealthy, it is important that the infected model maintains an accuracy similar to that of the benign model. The attacker's strategy will be to fault all the ReLUs of the targeted layer, while at the same time selecting a fraction, represented by $\chi$, of the training samples from the chosen target class to be subjected to faults. Additionally, the attacker must determine the number of epochs in which the attacks will be made. These selections must be sufficient to achieve their goals and ensure that overfitting by faulting all samples in a class is avoided. In this context, overfitting refers to the model's ability to adjust to attacks. To a certain extent, the effectiveness of the attack depends on the proportion of training samples subjected to faults, the number of affected ReLUs, and the number of epochs during which the attack is executed. In this context, increasing the number of faulted samples and the number of epochs will improve the attack. Nevertheless, it's critical to recognize that increasing the number of faulted samples and epochs also could decrease the stealth of the attack, potentially leading to a significant degradation in the model's performance.

\begin{figure*}[h]
\centering
\includegraphics[scale=0.14]{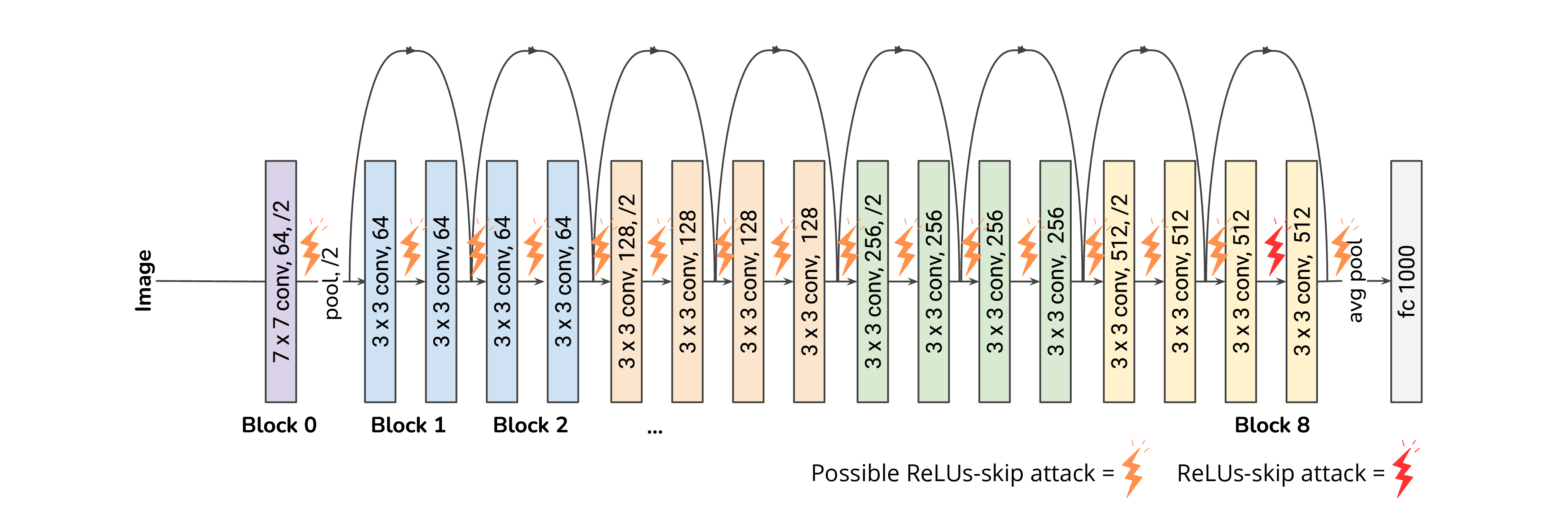}
\caption{\label{fig:img_faulting_strategy} Faulting strategy. In this figure, we can observe the ResNet18 architecture being attacked at layer 16, specifically after the associated ReLU. Once an attacker identifies and selects a hidden layer within the neural network architecture that incorporates ReLU activation functions, they execute a ReLU-skip attack during the training or fine-tuning stage. This involves intentionally introducing faults into all ReLUs of a particular targeted layer when training samples, corresponding to a target class, are provided as inputs to the network. Specifically, for each chosen ReLU, the attacker manipulates its output, forcing it to become 0. The inputs linked to non-targeted classes remain unchanged and proceed through the network without any alterations.}
\end{figure*}

\subsection{Fooling images generation strategy}
\label{sec:foolgenstrategy}
Notice that the \textit{faulting strategy} targets the training phase. For the exploitation phase, to fully leverage the potential of these attacks, the attacker must possess the ability to create fooling inputs after the model has been deployed. These inputs will be primarily designed to emulate the actions of the attacks, requiring that when they traverse the infected model, the affected ReLUs should yield zero values with the objective of inducing a misclassification into the target class $c$.

To generate the fooling inputs, following the assumptions outlined in Section \ref{sec:faultassu} for \approach{}, we propose that this process can be achieved by optimizing a custom loss function tailored for images. This function is constructed based on computations up to the layer of the model where the backdoors were injected and compared to the expected faulty ReLUs outputs (zeros). These computations within the model are performed on a specific fragment, which we denote as model~$\mathcal{M}'|_{1}^{\lambda}$ or, for simplicity, $\mathcal{M}_{\lambda}'$, since the attacker always need to feed the network from the first layer, so the subscript is not needed. Here, $\lambda$ represents the index of the compromised layer with the associated ReLU.

The optimization process is carried out with respect to the input of $\mathcal{M}_{\lambda}'$. The input of the model $\mathcal{M}_{\lambda}'$ is $\mathbf{x}'$, which is initialized as a copy of $\mathbf{x}$, so $\mathbf{x}'$ will change through the iterations of the optimization model until the confidence level exceeds a threshold $\tau$ or until a maximum number of iterations $\mu$ is reached. The threshold $\tau$ is the value that the confidence level should reach for a successful attack. Our custom cost function, which we will denote as $\mathcal{L}_{fool}$, consists of two complementary functions. The first, designated as $\mathcal{L}_{out}$, corresponds to the Huber Loss with the target value set to zero and the reduction operation set to the sum. This loss function is defined as 
\begin{equation}
    \mathcal{L\,}_{out} =
        \begin{cases}
            0.5 \cdot (x_n - y_n)^2, & \text{if  } |x_n - y_n| < \delta \\
            \delta \cdot (|x_n - y_n| - 0.5 \cdot \delta), & \text{otherwise.} \\
        \end{cases}\\
\end{equation}
In particular, we empirically set $\delta$ to 0.5, and due to the attack structure, the target is 0, thus $y_n=0$. $\mathcal{L\,}_{out}$ allows input adjustments to ensure that the compromised ReLUs produce zeros as intended.

On the other hand, the second function, $\mathcal{L\,}_{ssim}$, relates to the structure similarity index $SSIM$, which measures the similarity between two images. Therefore, $\mathcal{L\,}_{ssim}$  is outlined as  

\begin{equation}
    \mathcal{L\,}_{ssim} = 1 - SSIM(\mathbf{x}',\mathbf{x}).  
\end{equation}
$\mathcal{L\,}_{ssim}$ aims to keep the manipulated inputs as close to the base pattern $\mathbf{x}$ as possible. 

Therefore, our custom loss function for generating fooling images is the sum of $\mathcal{L\,}_{out} $ and $\mathcal{L\,}_{ssim}$, so 
\begin{equation}
    \mathcal{L}_{fool} = \mathcal{L\,}_{out} + \mathcal{L\,}_{ssim}.
    \label{eq:lossfunction}
\end{equation}
This provides a means to control the amount of change applied to the input $\mathbf{x}'$. Initially, $\mathcal{L\,}_{out}$ and $\mathcal{L\,}_{ssim}$ are on different scales. $\mathcal{L\,}_{ssim}$ starts at 0, since $\mathbf{x}$ and $\mathbf{x}'$ are identical, and then remains very close to 0. Conversely, $\mathcal{L\,}_{out}$ represents a type of error relative to the target 0. However, as optimization progresses, $\mathcal{L\,}_{out}$ converges to a similar scale. This characteristic allows $\mathcal{L}$ to primarily focus on exploiting the injected vulnerability.

In summary, $\mathcal{L}_{fool}$ serves a dual purpose: it enables the exploitation of the injected vulnerability while ensuring that $\mathbf{x}'$ closely resembles $\mathbf{x}$ in human perception.

\section{Evaluation}
\label{sec:evaluation}
Since our approach focuses on simulating physical \textit{ReLU-skip attacks}, we designed an implementation in code to replicate the desired behavior of this attack type. Subsequently, we apply the \textit{fooling image generation strategy} described in \Cref{sec:foolgenstrategy}. Finally, to evaluate the effectiveness of our approach, we report two of the attack metrics mentioned in \Cref{sec:attackmetric}, \ASR{}, and \ASRa{}. Following the evaluation protocol of previous work in the field~\cite{Wang_2023_CVPR}, the results are presented as the average of the \ASR{} of three target classes from the ImageNet dataset: Great Grey Owl, Goose, and French Bulldog. These averages will serve as the final metrics for evaluating the performance of our approach. It should be noted that some metrics depend on a confidence threshold $\tau$. For our evaluation, we empirically set $\tau$ to 0.1. A confidence value of 0.1 for the threshold $\tau$ means that the model assigns more than $10\%$ of probability to the predicted class. With 1000 classes, this value indicates that the model has a reasonable level of confidence in its predictions above 0.001 when all classes have the same probability.

The efficacy of \approach{} is evaluated, as in~\cite{Wang_2023_CVPR}, on three widely-used network architectures, including VGG-19~\cite{simonyan2014very}, ResNet-50~\cite{he2016deep}, and DenseNet-121~\cite{huang2017densely}.

\subsection{Experimental Setup}
\label{sec:exp_setup}
In this work, we extend the training of models previously trained on the ImageNet~\cite{deng2009imagenet} dataset, which can be considered a form of fine-tuning with the same dataset. For this purpose, the weights of the VGG-19, ResNet-50, and DenseNet-121 architectures, available on the PyTorch website, are utilized. 

\textbf{Datasets:} Building upon previous work~\cite{Wang_2023_CVPR}, we utilize two datasets in our experiments: the ImageNet~\cite{deng2009imagenet} and the Paintings~\cite{Paintings} datasets. The provided training and validation sets of ImageNet are employed. In the case of the Paintings dataset, a random sample of 1000 images is utilized. 

\textbf{Data preprocessing:} A batch size of 64 was empirically determined to be optimal due to considerations of memory limitations. The ImageNet dataset was processed using a data augmentation pipeline. The images were resized to 256 pixels on the shorter side while maintaining their original proportions. Additionally, random horizontal flipping and center cropping were applied to achieve a final size of $224\times224$ pixels. Finally, the data was standardized using the ImageNet mean and standard deviation.

\textbf{Model training (fine-tuning):} In the fine-tuning process, the infected model was generated using cross-entropy as the loss function and SGD as the optimizer. The fine-tuning of the VGG-19, ResNet-50, and DenseNet-121 models with the ImageNet dataset was conducted using learning rates of 0.0001, 0.001, and 0.001, respectively, with a momentum of 0.9 and a weight decay of $5\times10^{-4}$. The cosine annealing schedule was incorporated as the learning rate. Two major experiments were conducted. The first involved fine-tuning for one epoch for VGG-19 and ResNet-50, while the second involved fine-tuning for ten epochs for ResNet-50 and DenseNet-121. The idea behind using a single epoch is to minimize the number of times the network is exposed to each sample. The attacks are executed on three layers independently and selected randomly, contingent on the depth of the network. Thus, there is an infected model for an \textit{early} layer, one infected model for a \textit{middle} layer, and one infected model for a \textit{deep} layer at the end of the network but not belonging to the head.

\begin{table}[t]
\centering
\caption{Attack Success Rates $(\%)$ of \approach{} after attacking different target models on domain data (\textbf{ImageNet}). The table shows the average \ASR{} and \ASRa{}($\tau=0.1$) over the three selected target classes. Furthermore, the shown values were aggregated by how deep was the randomly attacked layer: \textit{early}, \textit{middle}, and \textit{deep}.}
\label{tab:results_ASR}
\begin{NiceTabular}{c|c|c|c|c}
    \toprule
    \textbf{Model} & \pbox{20cm}{\textbf{Epochs}\\ \textbf{Finetuning}} & \textbf{Layer} & \textbf{\ASRa{}} & \textbf{\ASR{}} \\
    \midrule
    \multirow{3}[3]{*}{VGG-19}  & \multirow{3}[3]{*}{1} & early & $5.63$ & $6.94$ \\ \cmidrule(r){3-5}
    & & middle & $37.07$ & $45.96$\\ \cmidrule{3-5}
    & & deep & $97.06$ & $98.30$\\ 
    \cmidrule{1-5}
    \multirow{6}[7]{*}{RestNet-50}  & \multirow{3}[3]{*}{1} & early & $7.16$ & $7.18$ \\ \cmidrule(r){3-5}
    & & middle & $44.80$ & $46.39$\\ \cmidrule{3-5}
    & & deep & $46.46$ & $70.35$\\ 
    \cmidrule{2-5}
    & \multirow{3}[3]{*}{10} & early & $13.75$ & $13.89$ \\ \cmidrule(r){3-5}
    & & middle & $75.54$ & $77.87$\\ \cmidrule{3-5}
    & & deep & $90.72$ & $97.94$\\ 
    \cmidrule{1-5}
    \multirow{3}{*}{DenseNet-121}  & 1 & deep & $0.0$ & $3.80$ \\ \cmidrule(r){3-5}
    \cmidrule{2-5}
    & 10 & deep & $0.0$ & $88.94$ \\
    \bottomrule    
\end{NiceTabular}
\end{table}

\textbf{Fooling image generation:} At this point, we have the infected models. Then we need to exploit the vulnerability. Subsequently, the vulnerability must be exploited. To achieve this, the objective function, which needed to be minimized, was defined by Equation \ref{eq:lossfunction} and explained in Section \ref{sec:foolgenstrategy}. Adam was employed as the optimizer, with a learning rate of 0.015 for VGG-19 and 0.025 for the other networks. These learning rates were chosen experimentally. The maximum number of iterations $\mu$ was empirically determined to be 200. In this case, the model is only able to see the sample 200 times (queries).

\textbf{Machine setup:} The experiments were conducted on a computational setup configured with an Intel i5-11400F CPU operating at 2.6 GHz. The system was equipped with 16 GB of DDR4 RAM clocked at 3200 MHz. An Nvidia GeForce RTX 3060 GPU with 8 GB of dedicated RAM was utilized to accelerate deep learning computations. 

\subsection{Attack overview}
\label{sec:gen_exp_setup}
In order to assess the efficacy of the attacks on each of the three different architectures across the three classes, we employ the following attack strategy. First, a model is selected. Second, a target class is selected for the faulting process. Third, the layer to be attacked is chosen. Fourth, the attack is executed. In this instance,  we set the value of $\chi$ to $0.9$. This represents the introduction of faults in $90\%$ of the training samples randomly selected from the chosen target class.

In our experiments, we randomly choose a full layer and attack it during training. Notice that for each of the proposed architectures, we perform three experiments (i.e., we perform the attack on three different layers independently). For DenseNet-121, only the final ReLU is attacked. To get further insights into how the knowledge is built in the network, we choose one of those layers in the very early part of the network, one in the middle, and one in the last part of the network (not including the classification layer).

Once the process of fine-tuning while faulting the network is completed, the infected model is ready for the fooling image generation, as detailed in Section~\ref{sec:foolgenstrategy}. In this phase, only a single image is required. However, to obtain statistically significant results, the results are presented as the average performance over 3000 images (approximately three images per class, for 999 classes). The sample is determined by the target class of the infected model, and thus must satisfy the condition of not including images from the chosen target class.

\begin{table}[t]
    \centering
    \caption{Mean classification accuracy of the attacked models compared against the accuracy of the baselines models (benign model).}
    \label{tab:acc_models}
    \begin{NiceTabular}{c|c|c|c}
        \toprule
        \textbf{Model} & \pbox{20cm}{\textbf{Benign model}\\ \textbf{accuracy (\%)}} & \pbox{20cm}{\textbf{Epochs}\\ \textbf{Finetuning}} &\pbox{20cm}{\textbf{Infected model}\\ \textbf{accuracy (\%)}} \\
        \midrule
        VGG-19 & $ 72.38 $ & 1 & $72.33 {\scriptstyle \pm 0.086} $\\
        \midrule
        \multirow{2}[2]{*}{ResNet-50} & \multirow{2}[2]{*}{$ 76.13 $} & 1  & $74.47 {\scriptstyle \pm 0.101}$ \\
        \cmidrule{3-4}
         & & 10 &  $75.18 {\scriptstyle \pm 0.101}$\\
         \midrule
        \multirow{2}[2]{*}{DenseNet-121} & \multirow{2}[2]{*}{$ 74.43 $} & 1 & $73.46 {\scriptstyle \pm 0.091}$\\
        \cmidrule{3-4}
         & & 10  & $74.47 {\scriptstyle \pm 0.101}$\\
        \bottomrule
    \end{NiceTabular}
\end{table}

\section{Results}
\label{sec:results}

\subsection{ImageNet Dataset}
\label{sec:results_imagenet}
The results are presented in the two phases of \approach: the faulting attack and the image generation. In the \Cref{tab:results_ASR} we present the results for the three evaluated architectures on the data domain of the models (ImageNet). In this scenario, attackers and users share the same data domain. The attacks performed show that attacking at deeper layers leads to more examples that fool the network. Remarkably, we can see that \approach{} achieves a \ASR{} superior to $98\%$ after performing the attack for only one epoch, especially in the case of VGG-19. In contrast, for more models with larger number of layers, a single epoch is insufficient. For instance, in ResNet-50 we achieve an \ASR{} of $70.35\%$ after one epoch, but we achieve $82.25\%$ after 10 epochs. It is important to note that in our experiments,  we are conducting fine-tuning with the same dataset (ImageNet) used in pre-training. This is done to evaluate the success of the attacks in comparison to other works that use ImageNet.

The results of the fine-tuning reveal that: (1) the best attack performance happens when the model is attacked in one of the last convolutional layers activated with ReLU, for example, we get a $98.30\%$ in \ASR{} for one epoch in VGG-19 (2) our approach is potentially hard to caught in real deployments since the models are still highly accurate. For instance, in the VGG-19 experiments, the infected model achieved $73.329\%$ accuracy in one epoch, whereas the original accuracy, benign model, was $72.376\%$, as shown in Table~\ref{tab:acc_models}. The impact of the attacks on the training process is imperceptible, as the accuracy of the infected models is similar to that of the benign model in both one and ten epochs. Consequently, the attack itself would likely go unnoticed.

\subsection{Paintings Dataset}
In \Cref{tab:results_ASR_paintings}, we present the results using the Paintings dataset, which is not part of the original domain of the models. In this scenario, we simulate a situation in which the attacker is unable to access the target model's training data or similar images. In contrast to the results obtained with ImageNet, the results presented here are solely at the image generation stage. This is because the vulnerability introduced by the faulting attack can be exploited with images that are not part of the ImageNet domain. The results indicate that attacks conducted in deeper layers have the highest \ASR{},  a finding that is consistent with the results obtained using ImageNet. For instance, we achieved an ASR of $99.70\%$ in VGG-19, and $93.87\%$ in  DenseNet-121.

\begin{table}[t]
\centering
\caption{Attack Success Rates $(\%)$ of \approach{} after attacking different target models on out-domain data (\textbf{Paintings}). The table shows the average \ASR{} and \ASRa{}($\tau=0.1$) over the three selected target classes. Furthermore, the shown values were aggregated by how deep was the randomly attacked layer: \textit{early}, \textit{middle}, and \textit{deep}.}
\label{tab:results_ASR_paintings}
\begin{NiceTabular}{c|c|c|c|c}
    \toprule
    \textbf{Model} & \pbox{20cm}{\textbf{Epochs}\\ \textbf{Finetuning}} & \textbf{Layer} & \textbf{\ASRa{}} & \textbf{\ASR{}} \\
    \midrule
    \multirow{3}[3]{*}{VGG-19}  & \multirow{3}[3]{*}{1} & early & $33.10$ & $37.43$ \\ \cmidrule(r){3-5}
    & & middle & $51.70$ & $60.37$\\ \cmidrule{3-5}
    & & deep & $98.73$ & $99.70$\\ 
    \cmidrule{1-5}
    \multirow{6}[7]{*}{RestNet-50}  & \multirow{3}[3]{*}{1} & early & $6.30$ & $6.30$ \\ \cmidrule(r){3-5}
    & & middle & $48.37$ & $49.30$\\ \cmidrule{3-5}
    & & deep & $30.77$ & $58.00$\\ 
    \cmidrule{2-5}
    & \multirow{3}[3]{*}{10} & early & $12.30$ & $12.40$ \\ \cmidrule(r){3-5}
    & & middle & $83.90$ & $85.47$\\ \cmidrule{3-5}
    & & deep & $89.30$ & $97.90$\\ 
    \cmidrule{1-5}
    \multirow{3}{*}{DenseNet-121}  & 1 & deep & $0.0$ & $3.4$ \\ \cmidrule(r){3-5}
    \cmidrule{2-5}
    & 10 & deep & $0.0$ & $93.87$ \\
    \bottomrule    
\end{NiceTabular}
\end{table}

\section{Discussion}

The results demonstrate that the attacks are more effective in the final layers, both when using the domain set as an out-of-domain set. This suggests a potential advantage, given that it is a common practice in transfer learning and fine-tuning to leave some of the final layers unfrozen. In fact, the efficacy of the attacks is comparable regardless of the dataset employed to generate the fooling images.  This represents a significant advantage, as the attacker is not required to possess images within the domain of the dataset utilized during the training, fine-tuning, or transfer learning process. Furthermore, the attack on one epoch demonstrates that \approach{} is both efficient and powerful, as the \textit{ReLU-skip attack} does not have to be applied in numerous instances.

Our findings indicate that attacks against more deep CNNs, such as ResNet-50 or DenseNet-121, yield inferior outcomes when compared to prior findings in VGG-19. Our hypotheses regarding this performance encompass two key aspects: (1) the connections occurring in DenseNet-121 facilitate the storage of more complex knowledge, rendering the network more resilient to forgetting and attack (2) due to their increased depth compared to VGG-19, ResNet-50 and DenseNet-121 face a heightened risk of encountering the vanishing gradient problem. As indicated in \cite{lu2019dying}, a deep ReLU network will eventually die in probability as the network's depth approaches infinity. Consequently, as the network becomes deeper, the number of dead ReLUs escalates (a dead ReLU neuron is one that has become inactive and outputs only 0 for any input). ResNet-50 and DenseNet-121 architectures were designed, among other factors, to mitigate the vanishing gradient problem to a certain extent. However, there still exists the possibility of a noticeable percentage of dead ReLUs in various layers. Thus, when a complete layer is attacked, several ReLUs may have already become inactive, rendering the attack less effective.

Concerning the fooling images generation phase \Cref{sec:foolgenstrategy}, which entails the exploitation of the vulnerability, certain parameters can be adjusted to exploit the vulnerability optimally. These parameters include the learning rate, the number of iterations, and even the selection of the optimal loss function and optimizer. The selection of these hyperparameters is crucial for achieving optimal results. Following experimentation, a general configuration that proved effective for different target classes is presented in sections \Cref{sec:foolgenstrategy} and \Cref{sec:exp_setup}. \Cref{tab:losses_resnet50_gen} illustrates the loss functions that were trialed in the generation of the fooling images and the impact of these loss functions on the \ASR{}, particularly in the ResNet-50 infected models.

\begin{table}[h]
\centering
    \caption{Ablation study on different loss functions used in the fooling image generation phase to exploit the attacks in the fine-tuned ResNet-50 models in one epoch for \textbf{ImageNet}. The learning rate used in fine-tuning is 0.001. In the fooling image generation phase, the learning rate is set to 0.025 and a maximum of 200 iterations or queries are allowed.}
    \label{tab:losses_resnet50_gen}
\begin{NiceTabular}{c|c|c|c}
    \toprule
    \textbf{Loss function} & \textbf{Layer} & \textbf{\ASRa{}}& \textbf{\ASR{}}\\
    \midrule
    \multirow{3}[4]{*}{L1} & early & $0.10$ & $0.10$\\
    \cmidrule{2-4}
    & middle &  $30.48$ &$31.41$\\
    \cmidrule{2-4}
    & deep &  $20.74$ &$36.40$\\
    \midrule
    \multirow{3}[4]{*}{L2} & early & $6.42$ & $6.42$\\
    \cmidrule{2-4}
    & middle &  $44.59$ & $46.03$\\
    \cmidrule{2-4}
    & deep  & $45.32$ & $70.10$\\
    \midrule
    \multirow{3}[4]{*}{Huber} & early & $7.16$ & $7.18$\\
    \cmidrule{2-4}
    & middle & $44.80$ & $46.39$\\
    \cmidrule{2-4}
    & deep & $46.46$ & $70.35$\\
    \bottomrule
\end{NiceTabular}
\end{table}

Finally, we show in \Cref{tab:comparison_attacks} a comprehensive benchmarking of \approach{} against related attack types. Particularly, for the same scenario, we show our performance in comparison to weights-oriented backdoor~\cite{yao2024deepvenom} and targeted evasion~\cite{wang2023towards}. Overall, \approach{} achieves either a competitive or higher attack success rate while featuring advantages that render it more attractive and effective than other related attacks. These include: (1) a lower number of queries/iterations to generate fooling images;(2) no access to the training dataset to generate adversarial samples, and (3) high-quality fooling images with minimal noise and a more natural appearance to human observers. Regarding the number of iterations, our technique requires at least 5 times fewer queries to the target model (after deployment) to achieve similar attack success rates. On the other hand, one strong advantage of \approach{} is that it only requires one image (i.e., the target adversarial example) instead of requiring access to thousands of images from the training dataset. Thirdly, \approach{} generates adversarial examples where the trigger is imperceptible for humans. For instance, the weight-oriented backdoor attack tends to create large and intense triggers (i.e., colored pattern) taking $9.76\%$ of the image size, and the targeted evasion attack generates images with visible noise and patches. In contrast, \approach{} generates adversarial examples where the noise is low and in most cases imperceptible. 

\section{Countermeasures}
\label{sec:countermeasures}

\textbf{Adversarial Training}: This is a defensive mechanism against adversarial attacks. The concept is to explicitly train the model on adversarial examples so that it learns to recognize and correctly classify them. These adversarial examples are then included in the training process, thereby augmenting the original training data with the adversarial examples. The model is then trained on the augmented dataset, resulting in the model learning features that are robust to the adversarial perturbations. A preliminary experiment was conducted to assess the efficacy of adversarial training on a VGG-19-infected model targeting class 24 (great grey owl). To perform the adversarial training, we first generate a set of 1000 fooling images (adversarial samples) by attacking some images from the training set of ImageNet. Afterwards, we then fine-tune the model (i.e., VGG-19) including a set of adversarial examples (1000), in addition to a sample of 34,745 images from the ImageNet training set. Finally, we test the ASR on the test dataset to check if the adversarial training was effective. For VGG-19, we found that after re-training with this augmented dataset and testing with the set of roughly 3K fooling images, the \ASR{} drops to approximately 5.33\%.

\section{Related Work}
\label{sec:related}
\begin{figure}
    \centering
    \includegraphics[width=0.8\linewidth]{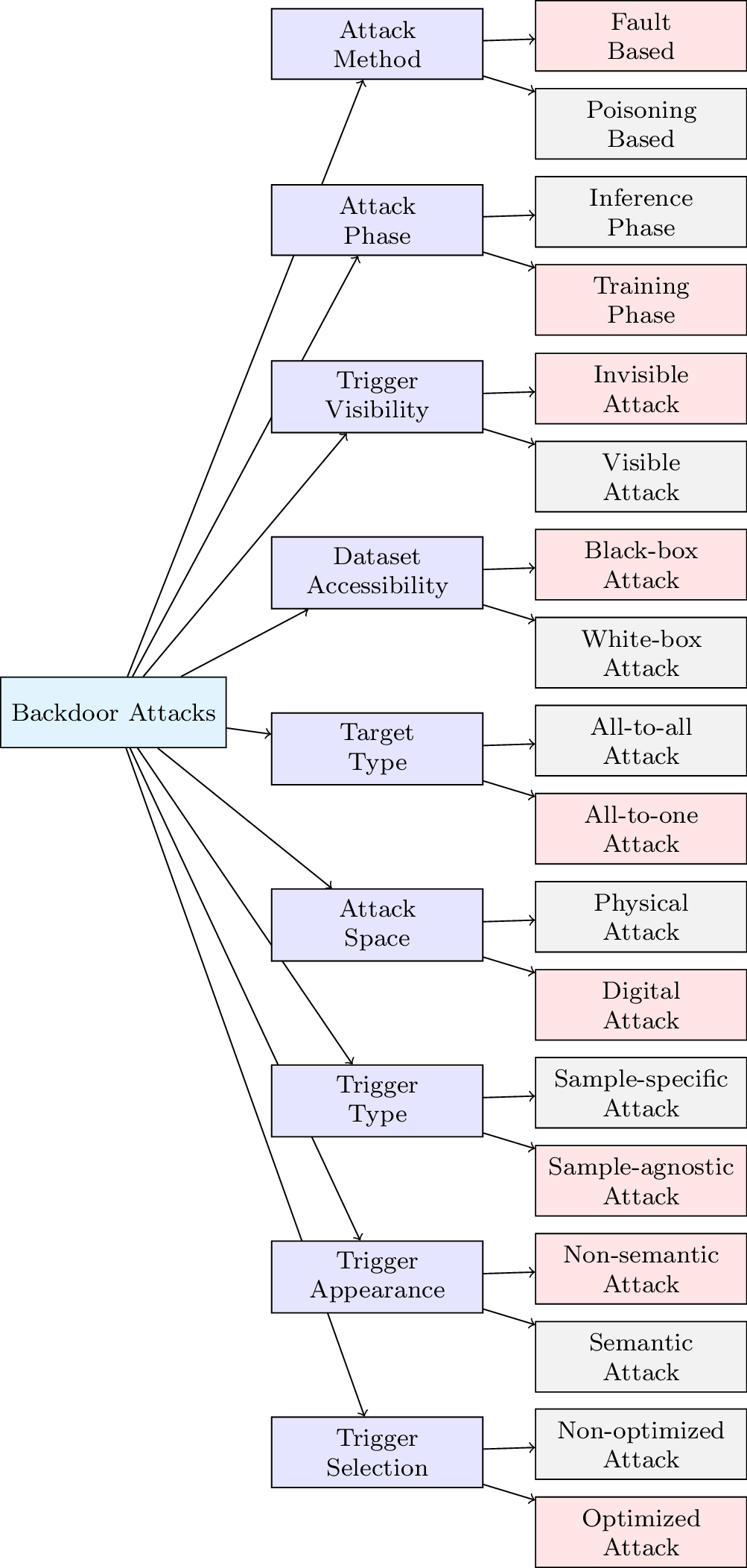}
    \caption{Classification of backdoor attacks inspired by~\cite{li2022backdoor}, with our \approach{} properties highlighted in red.}
    \label{fig:categorization}
\end{figure}

\begin{table*}[]
    \centering
    \caption{Benchmarking of \approach{} against state-of-the-art related attacks. For each attack type, we show the average attack success rate and the requirements needed to achieve this performance. We add an asterisk (*) to indicate that the comparison is not fully fair. For example, the ASR achieved by \cite{yao2024deepvenom} is computed on similar networks (i.e., VGG-16 and ResNet-18).}
    \label{tab:comparison_attacks}
    \begin{NiceTabular}{c | c |c| c | c} \toprule
        \multicolumn{2}{c |}{Attack type} & \textbf{Weights-oriented backdoor~\cite{yao2024deepvenom}} & \textbf{Targeted evasion~\cite{wang2023towards}} & \textbf{ReLU-skip} (this work) \\ \midrule
        & Tampers with & Training data & Model input & Training process \\
        \midrule
        \multirow{4}*{\textbf{ASR (\%)}}   & VGG-19 & 97.40 & 99.33 & 98.30\\
        \cmidrule{2-5}
                             & ResNet-50  & 97.00* & 99.16 & 97.94\\
        \cmidrule{2-5}
                             & DenseNet-121  & - & 99.16 & 88.94\\
        \midrule
        \multicolumn{2}{c |}{\# of model queries to generate fooling images}  & thousands & $>1000$ & 200\\ \midrule
        \multicolumn{2}{c |}{Access to training dataset}  & limited access ($5\%$) & full access & not required\\ \midrule
        \multicolumn{2}{c |}{Trigger visibility in the image} & High (visible colorful pattern) & Medium (visible noise) & Low (subtle noise) \\ \midrule
        \multicolumn{2}{c}{\multirow{9}{*}{Generated image examples}}& \multirow{9}{*}{\includegraphics[width=0.15\linewidth]{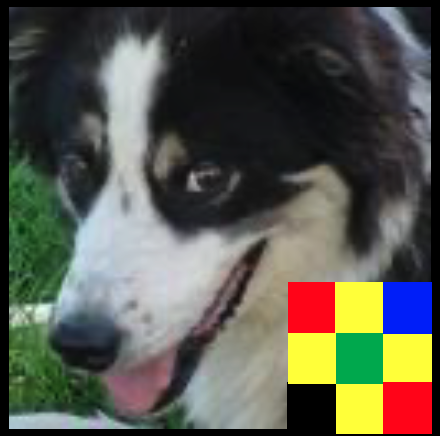}} & \multirow{9}{*}{\includegraphics[width=0.15\linewidth]{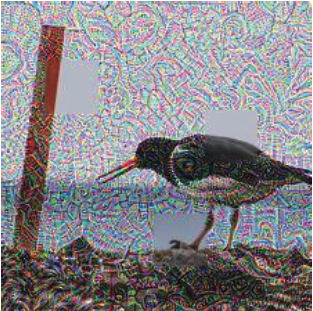}} & \multirow{9}{*}{\includegraphics[width=0.15\textwidth]{densenet121_fool_11_class_0_tclass_24.png}} \\
        \multicolumn{2}{c}{} & & & \\
        \multicolumn{2}{c}{} & & & \\
        \multicolumn{2}{c}{} & & & \\
        \multicolumn{2}{c}{} & & & \\
        \multicolumn{2}{c}{} & & & \\
        \multicolumn{2}{c}{} & & & \\
        \multicolumn{2}{c}{} & & & \\
        \multicolumn{2}{c}{} & & & \\
        \bottomrule
    \end{NiceTabular}
\end{table*}

\noindent
\textbf{Adversarial backdoor attacks.}
Backdoor attacks\footnote{In the literature, one can also find the term \textit{trojan} attack. These two terms can be used interchangeably.} on neural networks have been gaining attention in the adversarial learning community~\cite{trojan1,gu2017badnets,liu2018trojaning,zou2018potrojan}.
While the first works focused on adding additional hardware components to plant the backdoor~\cite{trojan1}, most of the follow-up works utilized changes in the training data to inject the backdoor~\cite{li2022backdoor} stealthily.
These attacks work by poisoning training samples to encode backdoor functionality during the training process~\cite{li2021backdoor,liu2020reflection}.
The backdoor is then activated by adding a trigger to the model input, which generally causes misclassification to a target class.
It was shown that this trigger can be hard to distinguish by the human eye~\cite{li2020invisible}.

\noindent
\textbf{Fault-based backdoor attacks.}
One of the challenges of the poisoning-based backdoor attack is that the adversary needs to be able to tamper with the training data, which might be impractical in many scenarios.
The answer to that are the fault-based backdoor attacks that modify the model parameters directly without changing the training data.
\begin{itemize}
    \item \textit{Inference phase attacks.} These attacks change the model weight after the training.
    The first attack in this direction presented a generic methodology for backdooring CNNs via targeted weight perturbations~\cite{dumford2020backdooring}. 
    As perturbing weights has been previously used for misclassifying outputs~\cite{rakin2019bit}, this hardware attack method was utilized by Rakin et al.~\cite{rakin2020tbt} to plant a backdoor at the bit-level.
    After that, several works adopted this method in various contexts~\cite{garg2020can,chen2021proflip}.
    Generally, weight-flipping attacks assume the attacker can use Rowhammer~\cite{mutlu2019rowhammer} -- a bug in random access memories that allows bit-flipping of target bits by rapidly rewriting the surrounding area of the memory.
    \item \textit{Training phase attacks.} There are currently two directions in this sub-category, each of them having a single representative as of now.
    The first direction is similar to the aforementioned attacks in the sense that the attacker utilizes Rowhammer to flip the bits -- but this time during the model training.
    DeepVenom is an instantiation of such an attack~\cite{yao2024deepvenom}.
    However, compared to DeepBaR, DeepVenom requires a visible large trigger image embedded in the original image and therefore is not imperceptible to human eyes.
    Our approach falls within the second direction -- tampering with the processor instructions during the training process.
    The closest work to ours is an attack called FooBaR~\cite{breier2022foobar} which inserts a backdoor during the training phase of the model by faulting activation functions.
    However, the generation of triggers for FooBaR requires constraint solving, which is generally an NP-complete problem.
    The solving complexity grows exponentially with the number of parameters, and therefore, does not scale to large networks that are used for image classification tasks such as CIFAR10 or ImageNet.
    Our approach, on the other hand, generates the fooling images with gradient descent and thus can be used on arbitrarily large networks.
    Another disadvantage of FooBar is that, unlike in DeepBaR, the fooling images are visibly different from the original ones.
\end{itemize}

To put our work into the context of backdoor attacks, we can use a classification chart shown in Figure~\ref{fig:categorization}, inspired by~\cite{li2022backdoor} but with two additional categories: attack method, and attack phase.
The attack method differentiates between the poisoning attacks, detailed at the beginning of this section, and fault-based attacks, where \approach{} belongs.
Because of that category, we had to add the attack phase as well -- while the poisoning attacks are always performed during the training, the bit-flip-based fault-based backdoors can also be injected during the inference.
In Figure~\ref{fig:categorization}, the categories where our attack belongs are highlighted in red color.


\noindent
\textbf{Defenses against backdoor attacks.}
There are two categories of defense methods against backdoor attacks: \textit{empirical defenses} and \textit{certified defenses}.
Empirical defenses~\cite{wang2019neural,kolouri2020universal,li2021anti} are based on characteristics of existing attacks and provide good effectiveness against those.
However, as there is no theoretical guarantee with this type, it is possible to adapt attacks in a way to bypass them~\cite{li2022backdoor}.
Certified defenses, on the other hand, provide provable security guarantees under some assumptions~\cite{wang2020certifying,xie2021crfl,weber2023rab}.
The performance of these is generally weaker compared to empirical defenses as it is hard to fulfill the given assumptions.

It is important to note that the existing defenses against backdoors protect against poisoning-based attacks, and therefore do not effectively protect against fault-based attacks~\cite{li2022backdoor}.

\textbf{Defenses against fault-based attacks.}
As the majority of fault-based attacks are weight-oriented, most of the defenses focus on protecting against bit flips in memory.
Techniques in this direction include checksums~\cite{li2021radar}, hashing~\cite{javaheripi2021hashtag}, encoding~\cite{velvcicky2024deepncode}, usage of binarized neural networks~\cite{rakin2021ra}, or in-DRAM swapping~\cite{zhou2023dnn}.
A method called NeuroPots~\cite{liu2023neuropots} utilizes a proactive honeypot approach to ``lure'' the attacker to flip certain bits in the network, making the detection and recovery more efficient.

To develop an appropriate countermeasure against ReLU-skip attacks, one would have to turn to techniques aimed at cryptographic implementations, as control-flow integrity protection is well-studied in this domain~\cite{werner2016protecting}.
One can use software methods such as instruction duplication/triplication to provide some level of protection~\cite{barenghi2010countermeasures} or replace a vulnerable instruction with a sequence of fault-tolerant ones~\cite{moro2014formal}.
In hardware, redundant circuits can be used to some extent -- however, it was shown that with precise equipment, one can simultaneously inject identical faults into these circuits~\cite{selmke2016attack}. 

Another class of countermeasures utilizes a separate device, usually a sensor~\cite{he2017fpga}, that can detect local disturbances of a potentially harmful character.
For a more complete overview of techniques used for protecting cryptographic implementations, an interested reader is suggested to look at~\cite{baksi2022survey}.

\section{Conclusions}

In this work, we introduce a novel fault-based backdoor attack, \approach{}, that allows attackers to trigger targeted misclassification for several image-based neural network architectures (such as ResNet-50, VGG-19, and DenseNet-121). Our technique relies on simple fault injection techniques that are applied during training (i.e., fine-tuning) stages. Overall, our approach achieves greater attack success rates than state-of-the-art targeted adversarial techniques without compromising the performance of the benign pre-trained model. Remarkably, our technique does not require full training of complex surrogate models to create further adversarial examples. Moreover, our approach requires 50 times fewer queries to the model than other backdoor attacks when creating adversarial examples. Additionally, we have presented a countermeasure to protect deployed systems against our proposed fault-based attacks.

\bibliographystyle{IEEEtran}
\bibliography{bibl}

\begin{thebibliography}{10}
\providecommand{\url}[1]{#1}
\csname url@samestyle\endcsname
\providecommand{\newblock}{\relax}
\providecommand{\bibinfo}[2]{#2}
\providecommand{\BIBentrySTDinterwordspacing}{\spaceskip=0pt\relax}
\providecommand{\BIBentryALTinterwordstretchfactor}{4}
\providecommand{\BIBentryALTinterwordspacing}{\spaceskip=\fontdimen2\font plus
\BIBentryALTinterwordstretchfactor\fontdimen3\font minus \fontdimen4\font\relax}
\providecommand{\BIBforeignlanguage}[2]{{%
\expandafter\ifx\csname l@#1\endcsname\relax
\typeout{** WARNING: IEEEtran.bst: No hyphenation pattern has been}%
\typeout{** loaded for the language `#1'. Using the pattern for}%
\typeout{** the default language instead.}%
\else
\language=\csname l@#1\endcsname
\fi
#2}}
\providecommand{\BIBdecl}{\relax}
\BIBdecl

\bibitem{jordan2015machine}
M.~I. Jordan and T.~M. Mitchell, ``Machine learning: Trends, perspectives, and prospects,'' \emph{Science}, vol. 349, no. 6245, pp. 255--260, 2015.

\bibitem{liu2020computing}
L.~Liu, S.~Lu, R.~Zhong, B.~Wu, Y.~Yao, Q.~Zhang, and W.~Shi, ``Computing systems for autonomous driving: State of the art and challenges,'' \emph{IEEE Internet of Things Journal}, vol.~8, no.~8, pp. 6469--6486, 2020.

\bibitem{shafaei2018uncertainty}
S.~Shafaei, S.~Kugele, M.~H. Osman, and A.~Knoll, ``Uncertainty in machine learning: A safety perspective on autonomous driving,'' in \emph{Computer Safety, Reliability, and Security: SAFECOMP 2018 Workshops, ASSURE, DECSoS, SASSUR, STRIVE, and WAISE, V{\"a}ster{\aa}s, Sweden, September 18, 2018, Proceedings 37}.\hskip 1em plus 0.5em minus 0.4em\relax Springer, 2018, pp. 458--464.

\bibitem{deng2020analysis}
Y.~Deng, X.~Zheng, T.~Zhang, C.~Chen, G.~Lou, and M.~Kim, ``An analysis of adversarial attacks and defenses on autonomous driving models,'' in \emph{2020 IEEE international conference on pervasive computing and communications (PerCom)}.\hskip 1em plus 0.5em minus 0.4em\relax IEEE, 2020, pp. 1--10.

\bibitem{lowd2005adversarial}
D.~Lowd and C.~Meek, ``Adversarial learning,'' in \emph{Proceedings of the eleventh ACM SIGKDD international conference on Knowledge discovery in data mining}, 2005, pp. 641--647.

\bibitem{ilyas2018black}
A.~Ilyas, L.~Engstrom, A.~Athalye, and J.~Lin, ``Black-box adversarial attacks with limited queries and information,'' in \emph{International conference on machine learning}.\hskip 1em plus 0.5em minus 0.4em\relax PMLR, 2018, pp. 2137--2146.

\bibitem{breier2022foobar}
J.~Breier, X.~Hou, M.~Ochoa, and J.~Solano, ``Foobar: Fault fooling backdoor attack on neural network training,'' \emph{IEEE Transactions on Dependable and Secure Computing}, 2022.

\bibitem{breier2022practical}
J.~Breier and X.~Hou, ``How practical are fault injection attacks, really?'' \emph{IEEE Access}, vol.~10, pp. 113\,122--113\,130, 2022.

\bibitem{simonyan2014very}
K.~Simonyan and A.~Zisserman, ``Very deep convolutional networks for large-scale image recognition,'' \emph{arXiv preprint arXiv:1409.1556}, 2014.

\bibitem{he2016deep}
K.~He, X.~Zhang, S.~Ren, and J.~Sun, ``Deep residual learning for image recognition,'' in \emph{Proceedings of the IEEE conference on computer vision and pattern recognition}, 2016, pp. 770--778.

\bibitem{huang2017densely}
G.~Huang, Z.~Liu, L.~Van Der~Maaten, and K.~Q. Weinberger, ``Densely connected convolutional networks,'' in \emph{Proceedings of the IEEE conference on computer vision and pattern recognition}, 2017, pp. 4700--4708.

\bibitem{weng2023comparative}
J.~Weng, Z.~Luo, D.~Lin, and S.~Li, ``Comparative evaluation of recent universal adversarial perturbations in image classification,'' \emph{Computers \& Security}, p. 103576, 2023.

\bibitem{yao2024deepvenom}
F.~Yao, ``Deepvenom: Persistent dnn backdoors exploiting transient weight perturbations in memories,'' in \emph{2024 IEEE Symposium on Security and Privacy (SP)}.\hskip 1em plus 0.5em minus 0.4em\relax IEEE Computer Society, 2024, pp. 244--244.

\bibitem{agarap2018deep}
A.~F. Agarap, ``Deep learning using rectified linear units (relu),'' \emph{arXiv preprint arXiv:1803.08375}, 2018.

\bibitem{gurney2018introduction}
K.~Gurney, \emph{An introduction to neural networks}.\hskip 1em plus 0.5em minus 0.4em\relax CRC press, 2018.

\bibitem{albawi2017understanding}
S.~Albawi, T.~A. Mohammed, and S.~Al-Zawi, ``Understanding of a convolutional neural network,'' in \emph{2017 international conference on engineering and technology (ICET)}.\hskip 1em plus 0.5em minus 0.4em\relax Ieee, 2017, pp. 1--6.

\bibitem{barenghi2012fault}
A.~Barenghi, L.~Breveglieri, I.~Koren, and D.~Naccache, ``Fault injection attacks on cryptographic devices: Theory, practice, and countermeasures,'' \emph{Proceedings of the IEEE}, vol. 100, no.~11, pp. 3056--3076, 2012.

\bibitem{kim2014flipping}
Y.~Kim, R.~Daly, J.~Kim, C.~Fallin, J.~H. Lee, D.~Lee, C.~Wilkerson, K.~Lai, and O.~Mutlu, ``Flipping bits in memory without accessing them: An experimental study of dram disturbance errors,'' \emph{ACM SIGARCH Computer Architecture News}, vol.~42, no.~3, pp. 361--372, 2014.

\bibitem{tang2017clkscrew}
A.~Tang, S.~Sethumadhavan, and S.~Stolfo, ``$\{$CLKSCREW$\}$: Exposing the perils of $\{$Security-Oblivious$\}$ energy management,'' in \emph{26th USENIX Security Symposium (USENIX Security 17)}, 2017, pp. 1057--1074.

\bibitem{qiu2019voltjockey}
P.~Qiu, D.~Wang, Y.~Lyu, and G.~Qu, ``Voltjockey: Breaching trustzone by software-controlled voltage manipulation over multi-core frequencies,'' in \emph{Proceedings of the 2019 ACM SIGSAC Conference on Computer and Communications Security}, 2019, pp. 195--209.

\bibitem{breier2015laser}
J.~Breier, D.~Jap, and C.-N. Chen, ``Laser profiling for the back-side fault attacks: with a practical laser skip instruction attack on aes,'' in \emph{Proceedings of the 1st ACM Workshop on Cyber-Physical System Security}, 2015, pp. 99--103.

\bibitem{kumar2017practical}
S.~D. Kumar, S.~Patranabis, J.~Breier, D.~Mukhopadhyay, S.~Bhasin, A.~Chattopadhyay, and A.~Baksi, ``A practical fault attack on arx-like ciphers with a case study on chacha20,'' in \emph{2017 Workshop on Fault Diagnosis and Tolerance in Cryptography (FDTC)}.\hskip 1em plus 0.5em minus 0.4em\relax IEEE, 2017, pp. 33--40.

\bibitem{liu2017fault}
Y.~Liu, L.~Wei, B.~Luo, and Q.~Xu, ``Fault injection attack on deep neural network,'' in \emph{Proceedings of the 36th International Conference on Computer-Aided Design}.\hskip 1em plus 0.5em minus 0.4em\relax IEEE Press, 2017, pp. 131--138.

\bibitem{breier2018practical}
J.~Breier, X.~Hou, D.~Jap, L.~Ma, S.~Bhasin, and Y.~Liu, ``Practical fault attack on deep neural networks,'' in \emph{Proceedings of the 2018 ACM SIGSAC Conference on Computer and Communications Security}, 2018, pp. 2204--2206.

\bibitem{hong2019terminal}
S.~Hong, P.~Frigo, Y.~Kaya, C.~Giuffrida, and T.~Dumitraș, ``Terminal brain damage: Exposing the graceless degradation in deep neural networks under hardware fault attacks,'' in \emph{28th USENIX Security Symposium (USENIX Security 19)}, 2019, pp. 497--514.

\bibitem{he2020defending}
Z.~He, A.~S. Rakin, J.~Li, C.~Chakrabarti, and D.~Fan, ``Defending and harnessing the bit-flip based adversarial weight attack,'' in \emph{Proceedings of the IEEE/CVF Conference on Computer Vision and Pattern Recognition}, 2020, pp. 14\,095--14\,103.

\bibitem{rakin2021t}
A.~S. Rakin, Z.~He, J.~Li, F.~Yao, C.~Chakrabarti, and D.~Fan, ``T-bfa: Targeted bit-flip adversarial weight attack,'' \emph{IEEE Transactions on Pattern Analysis and Machine Intelligence}, vol.~44, no.~11, pp. 7928--7939, 2021.

\bibitem{bai2021targeted}
J.~Bai, B.~Wu, Y.~Zhang, Y.~Li, Z.~Li, and S.-T. Xia, ``Targeted attack against deep neural networks via flipping limited weight bits,'' \emph{arXiv preprint arXiv:2102.10496}, 2021.

\bibitem{breier2021sniff}
J.~Breier, D.~Jap, X.~Hou, S.~Bhasin, and Y.~Liu, ``Sniff: reverse engineering of neural networks with fault attacks,'' \emph{IEEE Transactions on Reliability}, vol.~71, no.~4, pp. 1527--1539, 2021.

\bibitem{rakin2022deepsteal}
A.~S. Rakin, M.~H.~I. Chowdhuryy, F.~Yao, and D.~Fan, ``Deepsteal: Advanced model extractions leveraging efficient weight stealing in memories,'' in \emph{2022 IEEE Symposium on Security and Privacy (SP)}.\hskip 1em plus 0.5em minus 0.4em\relax IEEE, 2022, pp. 1157--1174.

\bibitem{rakin2020tbt}
A.~S. Rakin, Z.~He, and D.~Fan, ``Tbt: Targeted neural network attack with bit trojan,'' in \emph{Proceedings of the IEEE/CVF Conference on Computer Vision and Pattern Recognition}, 2020, pp. 13\,198--13\,207.

\bibitem{gross2023fpganeedle}
M.~Gross, J.~Krautter, D.~Gnad, M.~Gruber, G.~Sigl, and M.~Tahoori, ``Fpganeedle: Precise remote fault attacks from fpga to cpu,'' in \emph{Proceedings of the 28th Asia and South Pacific Design Automation Conference}, 2023, pp. 358--364.

\bibitem{courbon2015combining}
F.~Courbon, J.~J. Fournier, P.~Loubet-Moundi, and A.~Tria, ``Combining image processing and laser fault injections for characterizing a hardware aes,'' \emph{IEEE transactions on computer-aided design of integrated circuits and systems}, vol.~34, no.~6, pp. 928--936, 2015.

\bibitem{mutlu2019rowhammer}
O.~Mutlu and J.~S. Kim, ``Rowhammer: A retrospective,'' \emph{IEEE Transactions on Computer-Aided Design of Integrated Circuits and Systems}, vol.~39, no.~8, pp. 1555--1571, 2019.

\bibitem{wang2004image}
Z.~Wang, A.~C. Bovik, H.~R. Sheikh, and E.~P. Simoncelli, ``Image quality assessment: from error visibility to structural similarity,'' \emph{IEEE transactions on image processing}, vol.~13, no.~4, pp. 600--612, 2004.

\bibitem{Wang_2023_CVPR}
Z.~Wang, H.~Yang, Y.~Feng, P.~Sun, H.~Guo, Z.~Zhang, and K.~Ren, ``Towards transferable targeted adversarial examples,'' in \emph{Proceedings of the IEEE/CVF Conference on Computer Vision and Pattern Recognition (CVPR)}, June 2023, pp. 20\,534--20\,543.

\bibitem{deng2009imagenet}
J.~Deng, W.~Dong, R.~Socher, L.-J. Li, K.~Li, and L.~Fei-Fei, ``Imagenet: A large-scale hierarchical image database,'' in \emph{2009 IEEE conference on computer vision and pattern recognition}.\hskip 1em plus 0.5em minus 0.4em\relax Ieee, 2009, pp. 248--255.

\bibitem{Paintings}
``Painter by number,'' \url{https://www.kaggle.com/c/painter-by-numbers/data}, kaggle, 2017.

\bibitem{lu2019dying}
L.~Lu, Y.~Shin, Y.~Su, and G.~E. Karniadakis, ``Dying relu and initialization: Theory and numerical examples,'' \emph{arXiv preprint arXiv:1903.06733}, 2019.

\bibitem{wang2023towards}
Z.~Wang, H.~Yang, Y.~Feng, P.~Sun, H.~Guo, Z.~Zhang, and K.~Ren, ``Towards transferable targeted adversarial examples,'' in \emph{Proceedings of the IEEE/CVF Conference on Computer Vision and Pattern Recognition}, 2023, pp. 20\,534--20\,543.

\bibitem{li2022backdoor}
Y.~Li, Y.~Jiang, Z.~Li, and S.-T. Xia, ``Backdoor learning: A survey,'' \emph{IEEE Transactions on Neural Networks and Learning Systems}, 2022.

\bibitem{trojan1}
J.~Clements and Y.~Lao, ``Hardware trojan design on neural networks,'' in \emph{2019 IEEE International Symposium on Circuits and Systems (ISCAS)}, 2019, pp. 1--5.

\bibitem{gu2017badnets}
T.~Gu, B.~Dolan-Gavitt, and S.~Garg, ``Badnets: Identifying vulnerabilities in the machine learning model supply chain,'' \emph{arXiv preprint arXiv:1708.06733}, 2017.

\bibitem{liu2018trojaning}
Y.~Liu, S.~Ma, Y.~Aafer, W.-C. Lee, J.~Zhai, W.~Wang, and X.~Zhang, ``Trojaning attack on neural networks,'' in \emph{25th Annual Network And Distributed System Security Symposium (NDSS 2018)}.\hskip 1em plus 0.5em minus 0.4em\relax Internet Soc, 2018.

\bibitem{zou2018potrojan}
M.~Zou, Y.~Shi, C.~Wang, F.~Li, W.~Song, and Y.~Wang, ``Potrojan: powerful neural-level trojan designs in deep learning models,'' \emph{arXiv preprint arXiv:1802.03043}, 2018.

\bibitem{li2021backdoor}
Y.~Li, T.~Zhai, Y.~Jiang, Z.~Li, and S.-T. Xia, ``Backdoor attack in the physical world,'' \emph{arXiv preprint arXiv:2104.02361}, 2021.

\bibitem{liu2020reflection}
Y.~Liu, X.~Ma, J.~Bailey, and F.~Lu, ``Reflection backdoor: A natural backdoor attack on deep neural networks,'' in \emph{Computer Vision--ECCV 2020: 16th European Conference, Glasgow, UK, August 23--28, 2020, Proceedings, Part X 16}.\hskip 1em plus 0.5em minus 0.4em\relax Springer, 2020, pp. 182--199.

\bibitem{li2020invisible}
S.~Li, M.~Xue, B.~Z.~H. Zhao, H.~Zhu, and X.~Zhang, ``Invisible backdoor attacks on deep neural networks via steganography and regularization,'' \emph{IEEE Transactions on Dependable and Secure Computing}, vol.~18, no.~5, pp. 2088--2105, 2020.

\bibitem{dumford2020backdooring}
J.~Dumford and W.~Scheirer, ``Backdooring convolutional neural networks via targeted weight perturbations,'' in \emph{2020 IEEE International Joint Conference on Biometrics (IJCB)}.\hskip 1em plus 0.5em minus 0.4em\relax IEEE, 2020, pp. 1--9.

\bibitem{rakin2019bit}
A.~S. Rakin, Z.~He, and D.~Fan, ``Bit-flip attack: Crushing neural network with progressive bit search,'' in \emph{Proceedings of the IEEE/CVF International Conference on Computer Vision}, 2019, pp. 1211--1220.

\bibitem{garg2020can}
S.~Garg, A.~Kumar, V.~Goel, and Y.~Liang, ``Can adversarial weight perturbations inject neural backdoors,'' in \emph{Proceedings of the 29th ACM International Conference on Information \& Knowledge Management}, 2020, pp. 2029--2032.

\bibitem{chen2021proflip}
H.~Chen, C.~Fu, J.~Zhao, and F.~Koushanfar, ``Proflip: Targeted trojan attack with progressive bit flips,'' in \emph{Proceedings of the IEEE/CVF International Conference on Computer Vision}, 2021, pp. 7718--7727.

\bibitem{wang2019neural}
B.~Wang, Y.~Yao, S.~Shan, H.~Li, B.~Viswanath, H.~Zheng, and B.~Y. Zhao, ``Neural cleanse: Identifying and mitigating backdoor attacks in neural networks,'' in \emph{2019 IEEE Symposium on Security and Privacy (SP)}.\hskip 1em plus 0.5em minus 0.4em\relax IEEE, 2019, pp. 707--723.

\bibitem{kolouri2020universal}
S.~Kolouri, A.~Saha, H.~Pirsiavash, and H.~Hoffmann, ``Universal litmus patterns: Revealing backdoor attacks in cnns,'' in \emph{Proceedings of the IEEE/CVF Conference on Computer Vision and Pattern Recognition}, 2020, pp. 301--310.

\bibitem{li2021anti}
Y.~Li, X.~Lyu, N.~Koren, L.~Lyu, B.~Li, and X.~Ma, ``Anti-backdoor learning: Training clean models on poisoned data,'' \emph{Advances in Neural Information Processing Systems}, vol.~34, pp. 14\,900--14\,912, 2021.

\bibitem{wang2020certifying}
B.~Wang, X.~Cao, N.~Z. Gong \emph{et~al.}, ``On certifying robustness against backdoor attacks via randomized smoothing,'' \emph{arXiv preprint arXiv:2002.11750}, 2020.

\bibitem{xie2021crfl}
C.~Xie, M.~Chen, P.-Y. Chen, and B.~Li, ``Crfl: Certifiably robust federated learning against backdoor attacks,'' in \emph{International Conference on Machine Learning}.\hskip 1em plus 0.5em minus 0.4em\relax PMLR, 2021, pp. 11\,372--11\,382.

\bibitem{weber2023rab}
M.~Weber, X.~Xu, B.~Karla{\v{s}}, C.~Zhang, and B.~Li, ``Rab: Provable robustness against backdoor attacks,'' in \emph{2023 IEEE Symposium on Security and Privacy (SP)}.\hskip 1em plus 0.5em minus 0.4em\relax IEEE, 2023, pp. 1311--1328.

\bibitem{li2021radar}
J.~Li, A.~S. Rakin, Z.~He, D.~Fan, and C.~Chakrabarti, ``Radar: Run-time adversarial weight attack detection and accuracy recovery,'' in \emph{2021 Design, Automation \& Test in Europe Conference \& Exhibition (DATE)}.\hskip 1em plus 0.5em minus 0.4em\relax IEEE, 2021, pp. 790--795.

\bibitem{javaheripi2021hashtag}
M.~Javaheripi and F.~Koushanfar, ``Hashtag: Hash signatures for online detection of fault-injection attacks on deep neural networks,'' in \emph{2021 IEEE/ACM International Conference On Computer Aided Design (ICCAD)}.\hskip 1em plus 0.5em minus 0.4em\relax IEEE, 2021, pp. 1--9.

\bibitem{velvcicky2024deepncode}
P.~Vel{\v{c}}ick{\`y}, J.~Breier, M.~Kova{\v{c}}evi{\'c}, and X.~Hou, ``Deepncode: Encoding-based protection against bit-flip attacks on neural networks,'' \emph{arXiv preprint arXiv:2405.13891}, 2024.

\bibitem{rakin2021ra}
A.~S. Rakin, L.~Yang, J.~Li, F.~Yao, C.~Chakrabarti, Y.~Cao, J.-s. Seo, and D.~Fan, ``{RA-BNN: Constructing robust \& accurate binary neural network to simultaneously defend adversarial bit-flip attack and improve accuracy},'' \emph{arXiv preprint arXiv:2103.13813}, 2021.

\bibitem{zhou2023dnn}
R.~Zhou, S.~Ahmed, A.~S. Rakin, and S.~Angizi, ``{DNN}-defender: An in-{DRAM} deep neural network defense mechanism for adversarial weight attack,'' \emph{arXiv preprint arXiv:2305.08034}, 2023.

\bibitem{liu2023neuropots}
Q.~Liu, J.~Yin, W.~Wen, C.~Yang, and S.~Sha, ``$\{$NeuroPots$\}$: Realtime proactive defense against $\{$Bit-Flip$\}$ attacks in neural networks,'' in \emph{32nd USENIX Security Symposium (USENIX Security 23)}, 2023, pp. 6347--6364.

\bibitem{werner2016protecting}
M.~Werner, E.~Wenger, and S.~Mangard, ``Protecting the control flow of embedded processors against fault attacks,'' in \emph{Smart Card Research and Advanced Applications: 14th International Conference, CARDIS 2015, Bochum, Germany, November 4-6, 2015. Revised Selected Papers 14}.\hskip 1em plus 0.5em minus 0.4em\relax Springer, 2016, pp. 161--176.

\bibitem{barenghi2010countermeasures}
A.~Barenghi, L.~Breveglieri, I.~Koren, G.~Pelosi, and F.~Regazzoni, ``Countermeasures against fault attacks on software implemented aes: effectiveness and cost,'' in \emph{Proceedings of the 5th Workshop on Embedded Systems Security}, 2010, pp. 1--10.

\bibitem{moro2014formal}
N.~Moro, K.~Heydemann, E.~Encrenaz, and B.~Robisson, ``Formal verification of a software countermeasure against instruction skip attacks,'' \emph{Journal of Cryptographic Engineering}, vol.~4, pp. 145--156, 2014.

\bibitem{selmke2016attack}
B.~Selmke, J.~Heyszl, and G.~Sigl, ``Attack on a dfa protected aes by simultaneous laser fault injections,'' in \emph{2016 Workshop on Fault Diagnosis and Tolerance in Cryptography (FDTC)}.\hskip 1em plus 0.5em minus 0.4em\relax IEEE, 2016, pp. 36--46.

\bibitem{he2017fpga}
W.~He, J.~Breier, S.~Bhasin, N.~Miura, and M.~Nagata, ``An fpga-compatible pll-based sensor against fault injection attack,'' in \emph{2017 22nd Asia and South Pacific Design Automation Conference (ASP-DAC)}.\hskip 1em plus 0.5em minus 0.4em\relax IEEE, 2017, pp. 39--40.

\bibitem{baksi2022survey}
A.~Baksi, S.~Bhasin, J.~Breier, D.~Jap, and D.~Saha, ``A survey on fault attacks on symmetric key cryptosystems,'' \emph{ACM Computing Surveys}, vol.~55, no.~4, pp. 1--34, 2022.

\end{thebibliography}

\end{document}